\documentclass{article}

\usepackage{microtype}
\usepackage{graphicx}
\usepackage{subcaption}
\usepackage{booktabs} 

\usepackage{hyperref}

\usepackage[preprint]{icml2026/icml2026}

\usepackage{amsmath}
\usepackage{amssymb}
\usepackage{mathtools}
\usepackage{amsthm}
\usepackage{multirow}
\usepackage{tikz}
\usetikzlibrary{positioning, fit, shapes, arrows.meta}

\usepackage[capitalize,noabbrev]{cleveref}

\theoremstyle{plain}

\theoremstyle{definition}

\theoremstyle{remark}

\usepackage[textsize=tiny]{todonotes}

\icmltitlerunning{PABU: Progress-Aware Belief Update for Efficient LLM Agents}

\usepackage{tcolorbox} 
\usepackage{listings}
\usepackage{xcolor}

\lstdefinelanguage{XML}{
  morestring=[b]",
  morecomment=[s]{<!--}{-->},
  morekeywords={goal,progress,last_observation,saved_observation,
    available_action,attempted_action,observation_update,
    progress_update,action_update},
  sensitive=false
}

\usepackage[acronym]{glossaries}
\makeglossaries

\newacronym{pabu}{PABU}{Progress-Aware Belief Update}
\newacronym{llm}{LLM}{Large Language Model}
\newacronym{rl}{RL}{reinforcement learning}
\newacronym{eto}{ETO}{Exploration-based Trajectory Optimization}
\newacronym{dpo}{DPO}{Direct Preference Optimization}
\newacronym{sft}{SFT}{Supervised Fine-Tuning}
\newacronym{gigpo}{GiGPO}{Group-in-Group Policy Optimization}
\newacronym{pacs}{PACS}{Progress-Aware Context Selection}
\newacronym{pomdp}{POMDP}{Partially Observable Markov Decision Processes}
\newacronym{sota}{SoTA}{State of the art}

\hypersetup{
    linkcolor=black,    
}

\definecolor{myblue}{RGB}{49,123,178} 

\newcommand{\blue}[1]{\textcolor{myblue}{#1}}

\begin{document}

\twocolumn[
  \icmltitle{PABU: Progress-Aware Belief Update for Efficient LLM Agents}



  \icmlsetsymbol{equal}{*}

  \begin{icmlauthorlist}
    \icmlauthor{Haitao Jiang}{ncsu}
    \icmlauthor{Lin Ge}{amazon}
    \icmlauthor{Hengrui Cai}{uci}
    \icmlauthor{Rui Song}{amazon}
  \end{icmlauthorlist}

    \icmlaffiliation{ncsu}{North Carolina State University, Raleigh, NC, USA}
    \icmlaffiliation{uci}{University of California, Irvine, CA, USA}
    \icmlaffiliation{amazon}{Amazon, Seattle, WA, USA (This work does not relate to the positions at Amazon.)}
    
  \icmlcorrespondingauthor{Rui Song}{songray@gmail.com}

  \icmlkeywords{Machine Learning, ICML}

  \vskip 0.3in
]

\begingroup
\renewcommand{\thefootnote}{}
\footnotetext{Implementation available at: \url{https://github.com/Hunter-Jiang/Progress-Aware-Belief-Update}.}
\endgroup



\printAffiliationsAndNotice{}  

\begin{abstract}
\gls{llm} agents commonly condition actions on full action--observation histories, which introduce task-irrelevant information that easily leads to redundant actions and higher inference cost.
We propose \gls{pabu}, a belief-state framework that compactly represents an agent's state by explicitly modeling task progress and selectively retaining past actions and observations.
At each step, the agent predicts its relative progress since the previous round and decides whether the newly encountered interaction should be stored, conditioning future decisions only on the retained subset.
Across eight environments in the AgentGym benchmark, and using identical training trajectories, \gls{pabu} achieves an 81.0\% task completion rate, outperforming previous \gls{sota} models with full-history belief by 23.9\%.
Additionally, \gls{pabu}'s progress-oriented action selection improves efficiency, reducing the average number of interaction steps to 9.5, corresponding to a 26.9\% reduction.
Ablation studies show that both explicit progress prediction and selective retention are necessary for robust belief learning and performance gains.
\end{abstract}

\begin{figure*}
    \centering
    \includegraphics[width=0.66\linewidth]{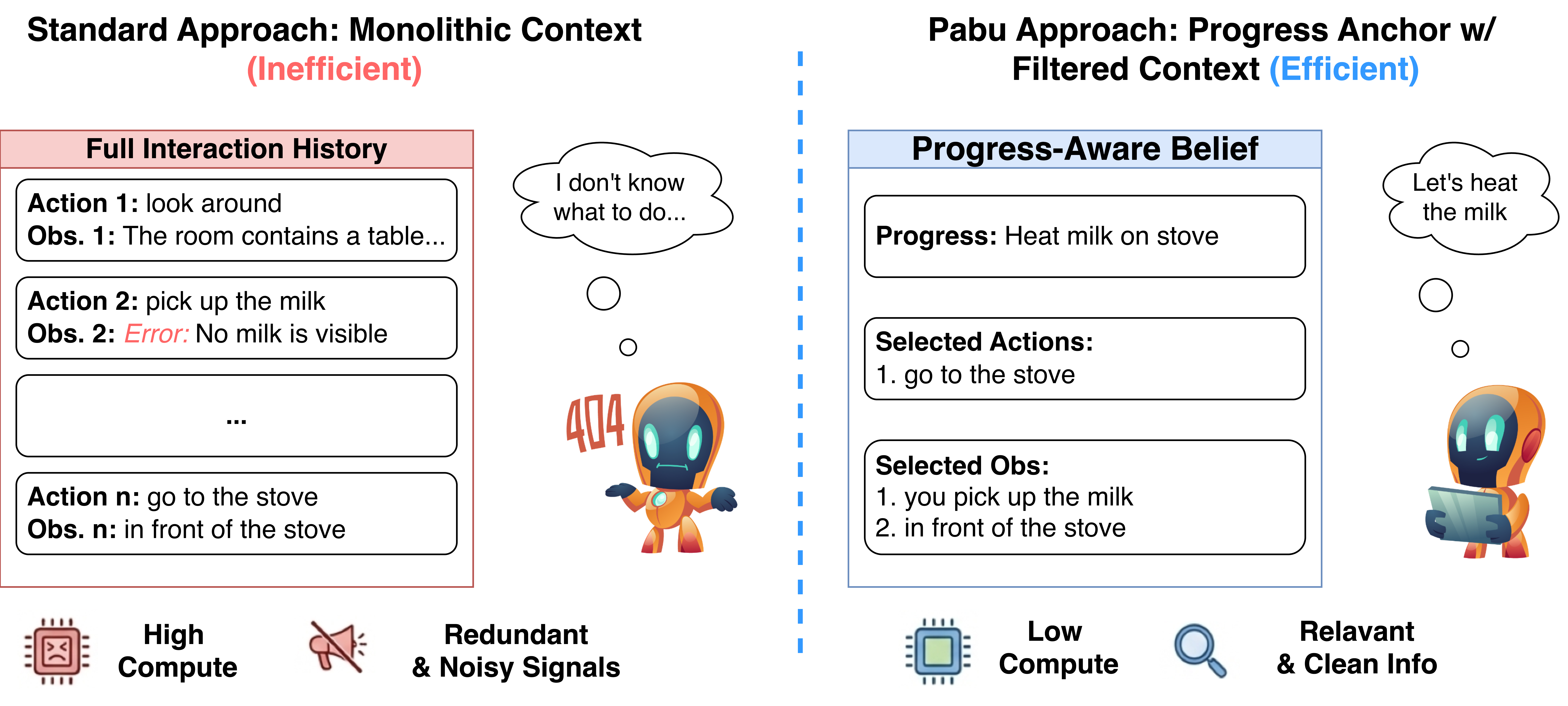}
    \caption{\textbf{Motivation for Progress-aware Belief Update.}
In a task such as heating and drinking milk, conventional methods rely on full interaction histories (left), which are noisy, redundant, and expensive to process. Our method (right) performs progress-aware belief updates, preserving only essential information and yielding a compact context that supports efficient and reliable learning.}
    \label{fig:motivation}
\end{figure*}

\section{Introduction}
Large Language Models (\glspl{llm}) are increasingly deployed as agents \citep{wei2026agenticreasoninglargelanguage} rather than merely as information extractors, leveraging capabilities such as self-reflection \citep{shinn2023reflexionlanguageagentsverbal}, retrieval-augmented generation \citep{gao2024retrievalaugmentedgenerationlargelanguage}, and tool invocation \citep{schick2023toolformerlanguagemodelsteach}.
In many real-world and simulated settings \citep{shridhar2020alfworld, yao2023webshopscalablerealworldweb}, \glspl{llm} must perform sequential actions while observing only the immediate outcomes of their interactions rather than the full environment state.
This partial observability poses a fundamental challenge: interaction history is widely used as a proxy for latent state \citep{ma2024agentboardanalyticalevaluationboard,xi2024agentgym}, yet because agentic data are typically collected under outcome-level supervision, where a single terminal success signal is assigned to the entire episode, such histories often contain weakly informative or task-irrelevant content. As a result, full-trajectory conditioning entangles state-relevant signals with spurious details \citep{chung2025evaluatinglongcontextreasoningllmbased}, increases computational overhead \citep{kang2025aconoptimizingcontextcompression}, and degrades overall performance \citep{202512.2119}. 
Effective state representation is therefore critical for enabling \gls{llm} agents to act efficiently and generalize across heterogeneous environments.
We argue that task progress provides a compact, task-agnostic signal for state tracking under partial observability.

Many agentic works focus primarily on aligning \glspl{llm} to perform effectively in interactive environments, rather than on learning compact state representations from interaction histories.
For example, frameworks such as \citet{qin2023toolllmfacilitatinglargelanguage} and \citet{zhang2024agentohanadesignunifieddata} unify instruction-following pipelines, while methods like Exploration-based Trajectory Optimization \citep{song2024trialerrorexplorationbasedtrajectory} and Group-in-Group Policy Optimization \citep{feng2025groupingrouppolicyoptimizationllm} optimize \gls{llm} behavior using success-failure trajectories or group-level action feedback via preference optimization.
AgentEvol \citep{xi2024agentgym} integrates \gls{rl} rewards directly into \gls{sft} loss, and \citet{chen2025atlasagenttuninglearning} shows that training on a carefully selected subset of interactions can further improve agent performance.
However, all these approaches treat the entire action–observation history as the agent’s state representation, which can lead to representation collapse when the history contains weakly informative or task-irrelevant actions. As a result, learning effective agent behavior becomes brittle, even when overall task success improves.

In traditional \gls{rl}, belief states \citep{rodriguez1999reinforcement, li2009multi} address partial observability by maintaining an estimate of the latent environment state.
However, such approaches typically rely on accurate transition or observation models \citep{Lauri_2023}, assumptions that are difficult to satisfy in complex, heterogeneous agentic settings.
In \glspl{llm}, prior work on context management has largely focused on summarization or retrieval of interaction histories, including compression-based summaries and segmented memory retrieval \citep{lu2025scalingllmmultiturnrl, kang2025aconoptimizingcontextcompression, lidayan2025abbelllmagentsacting, pan2025memoryconstructionretrievalpersonalized}.
While practical for controlling context length, these approaches assume that the interaction history itself, whether summarized or retrieved, constitutes a sufficient state representation.
This assumption introduces additional computational overhead and leaves the agent vulnerable to information loss, since summarization mechanisms learned from web-scale corpora may not faithfully preserve the task-relevant information required in heterogeneous agentic environments.

We aim to learn a compact belief state for long-horizon decision-making without relying on long-range summarization.
In our formulation, task progress is an estimated signal inferred from interaction trajectories, and belief updates are governed by a learned retention policy that decides whether newly observed actions and observations are stored.
Using progress as a task-agnostic signal, the belief state is modeled end-to-end and represents both estimated task advancement and a selectively retained subset of past interactions.
This allows \gls{llm}-based agents to operate over extended action-observation trajectories without treating the full interaction history as a monolithic context.
Rather than summarizing or retrieving past interactions, our approach performs progress-aware belief updates that retain only information necessary for next-step planning and control.
As a result, this belief-centric representation improves learning efficiency and decision accuracy while reducing inference-time context length and computational cost.

Our contributions are summarized as follows:
\begin{enumerate}
    \item We formalize \gls{pabu}, a belief-state framework for \gls{llm}-based agents that tracks task progress via a learned signal and selectively updates context through learned retention policies, eliminating the need for long-horizon episode summaries.

    \item We introduce an \gls{llm}-driven training objective that partitions actions into progress-consistent and non-progress-consistent groups with the progress signal. The policy is trained on all explored belief states to promote generalization, while actions are augmented toward progress-consistent ones for efficiency.

    \item We present {empirical results} on the \textit{AgentGym} suite.
    Agents trained with \gls{pabu} achieve an {81\% average task completion rate}, a {23.9\% improvement} over previous \gls{sota} baselines, while reducing the average number of interactions by {26.9\% (from 13.0 to 9.5 rounds)}.

    \item We conduct ablation analyses on model scale, retention loss, and training objectives to examine their influence on learned belief update behavior, providing insights into belief design and when selective action or observation retention and model scaling are most beneficial for agentic decision-making.
\end{enumerate}

\section{Problem Formulation} 
In this section, we formalize the agent–environment interaction process of agentic tasks as a \gls{pomdp} and introduce the belief-state abstraction underlying our method.

\subsection{Environment and Agentic Tasks} 
We consider agentic tasks in which an autonomous \gls{llm} iteratively respond to a user query by interacting with an external environment $\mathcal{E}$. An episode begins with a user query and proceeds over a finite horizon $H$. At each interaction step, the agent selects an action, such as a tool or API request, and receives a response from the environment. The agent iteratively reasons and acts until it emits a designated final-answer action or the interaction budget is exhausted.

\begin{figure}
    \centering
    \includegraphics[width=0.88\columnwidth]{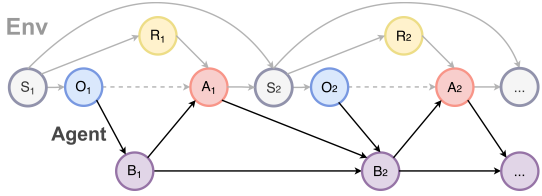}
    \caption{\textbf{POMDP Formulation with Belief State Estimation.} 
The \gls{llm} agent has no direct access to the latent \textcolor{gray}{environment states $S_\cdot$ or their dynamics}. 
Instead, it maintains a belief state $B_\cdot$ to estimate the current status from partial observations $O_\cdot$ and past actions $A_\cdot$, then uses this belief to guide action selection.}
    \label{fig:POMDP}
\end{figure}

Agentic tasks can be naturally modeled as a \textit{\gls{pomdp}} \citep{he2024wordsactionsunveilingtheoretical, zhang2026landscapeagenticreinforcementlearning}, defined by the six-tuple
$
\mathcal{M} = (\mathcal{S}, \mathcal{A}, \mathcal{O}, \mathcal{T}, O, R),
$
where $\mathcal{S}$ denotes the space of latent environment states, $\mathcal{A}$ is the action space, and $\mathcal{O}$ is the observation space consisting of textual feedback from the environment. The environment dynamics are governed by a transition function $\mathcal{T}(s' \mid s, a)$ and an observation model $O(o \mid s', a)$. The reward function $R(s)$ is defined only at terminal states and indicates task success.

At step $n$, the agent executes an action $a_n$, the environment transitions to a latent state $s_n \sim \mathcal{T}(\cdot \mid s_{n-1}, a_{n-1})$, and returns an observation $o_n \sim O(\cdot \mid s_n, a_{n-1})$. During inference, the agent has no access to 
$s_n$ or the reward signal.

\subsection{Belief State Abstraction}
To act effectively under partial observability, the agent maintains an internal \textit{Belief State} $b_n \in \mathcal{B}$ that summarizes the interaction history up to step $n$. Let $h_n = (q, a_1, o_1, \ldots, a_n, o_n)$ denote the user query $q$ together with the full action-observation history. In principle, there exists an optimal belief mapping $\Phi^*$ such that $b_n^* = \Phi^*(h_n)$ is a sufficient statistic for decision making and satisfies the Markov property with respect to the underlying \gls{pomdp}.

In practice, neither the environment dynamics nor the optimal belief mapping are known. Standard \gls{llm} agents therefore condition their actions directly on the full history $h_n$, encoded as a growing textual prompt, which introduces task-irrelevant information, increases inference cost, and often leads to redundant or suboptimal actions. 
We instead model belief update as a learned function
\begin{equation}
b_n = \Phi(b_{n-1}, a_{n-1}, o_{n-1}), \quad b_0 = q,  
\label{eqn:belief_iter}
\end{equation}
and action selection as
$a_n \sim \pi(a \mid b_n).$ where $\Phi$ produces a compact belief that conditions future decisions.
Both $\Phi$ and $\pi$ are realized by a single \gls{llm} $\pi_\theta$ with $\theta$ being trainable parameters, which jointly maintain beliefs and selects actions via generation. The belief state $b_n$ is a compact, text-based representation ($b_n \in \mathcal{B}$) that conditions future decisions, while the underlying distribution over latent states is learned implicitly through joint training on action trajectories.

\subsection{Optimization Objective}
Our objective is to maximize the expected terminal reward
\begin{equation}
\max_\theta \mathbb{E}_{\pi_\theta}\left[ R(s_t) \right],
\end{equation}
where $t$ denotes the termination step of the episode. Here, $\Phi_\theta$ constructs the belief state $b_n$ and $\pi_\theta$ selects actions $a_n \sim \pi_\theta(a \mid b_n)$, both learned jointly. Our goal is to learn a belief-state mapping $\hat{\Phi}$ that approximates the optimal $\Phi^*$ while retaining only task-relevant information from the 
history.

Building on this formulation, we introduce \gls{pabu}, a belief-state framework that explicitly models task progress and selectively retains past actions and observations. The resulting belief state provides a compact, structured summary of the agent’s state, enabling more efficient and effective decision making than conditioning on full histories.

\section{Progress-Aware Belief Update}
In this section, we introduce \gls{pabu}
, which maintains a compact belief state composed of two elements: an explicit estimate of task progress and a selectively retained subset of past actions and observations. This learned belief state is updated at each decision step and used autoregressively for action selection, enabling efficient and robust state tracking under partial observability.

\subsection{Progress as the Backbone of Belief State}
\gls{llm} agents operate in diverse and heterogeneous environments, making direct estimation of the underlying environment state, an assumption common in traditional \gls{rl}, both intractable and costly. The latent state space is typically unknown, high-dimensional, and only weakly reflected through textual observations.

Instead, we model agentic \textit{progress}, a text-based abstraction that captures the agent’s intentional advancement toward task completion. Progress is task-dependent but environment-agnostic: it represents what must be achieved rather than how it is realized in a specific environment. This abstraction provides a more tractable and generalizable anchor for belief construction than the latent environment state. Formally, let the progress associated with the action-observation pair at step $n$ be denoted by $p_n$. Progress evolves according to
\begin{equation}
p_{n+1} \sim \mathcal{T}_p(\cdot \mid p_n, a_n, o_n),
\end{equation}
where $\mathcal{T}_p$ models how task progress advances in response to the agent’s interactions. If an action does not complete the current task stage, the progress remains unchanged, i.e., $p_{n+1} = p_n$. Progress is maintained until the agent completes the current stage, at which point the next progress is predicted. At each step, the model estimates whether the current progress has been completed before selecting the next action.

This design yields progress sequences that are consistent and interpretable while remaining flexible across diverse tasks. Intuitively, the progress transition approximates the latent environment dynamics $(\mathcal{T}, O)$ projected onto explored task-relevant subsets. For example, in a household task, progress may correspond to locating an object, manipulating it, or verifying its status. Because progress sequences can be synthesized from successful trajectories using an \gls{llm}-based annotation process (our implementation described in Section \ref{sec:alg}), a custom-tuned \gls{llm} can model these transitions without direct access to the latent environment state, providing a stable backbone for belief-state construction under partial observability.

Concatenating completed progress stages yields a high-level summary of task execution. While progress does not strictly satisfy the Markov property, it serves as an approximately Markovian abstraction that substantially reduces dependence on long action-observation histories. Compared to raw action sequences, progress representations are shorter since multiple failed attempts may correspond to the same progress, and more semantically meaningful. Training on successful trajectories encourages consistent semantic clustering of progress descriptions, enabling the \gls{llm} to predict only the current progress rather than summarize the full episode history. This design significantly reduces inference costs compared to summarizing past interaction histories at every step or whenever they exceed a preset length.

\subsection{Selective Retention Mechanism}
\begin{figure}[t]
    \centering
    \includegraphics[width=0.7\linewidth]{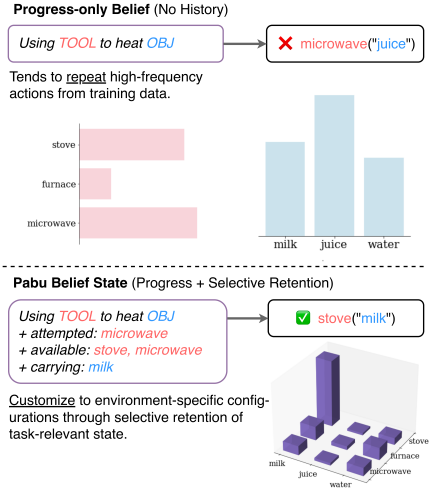}
    \caption{\textbf{\gls{pabu} Belief State Design.} Progress-only belief states are under-specified when identical progress stages require different actions across environments, while \gls{pabu} augments progress with selective retention of task-relevant actions and observations.}
    \label{fig:history_cond}
\end{figure}

While progress provides a stable backbone for belief construction, it is insufficient when multiple action attempts correspond to the same progress stage. Conditioning solely on progress can cause the agent to repeat failed actions or ignore environment-specific constraints as illustrated in Fig \ref{fig:history_cond}. To address this, \gls{pabu} augments progress with a \textit{selective retention mechanism} that preserves task-relevant actions and observations across steps, which comprises three components: fixed retention rules, progress-conditioned action attempts, and learned observation retention.

Certain elements are always retained due to their immediate relevance. The most recent observation is always included, as it provides the strongest signal for estimating whether the current progress has been completed, and the user query is retained throughout the episode as immutable task context.

When multiple actions are attempted under the same progress stage, the agent must remember which actions have already been tried. \gls{pabu} therefore maintains a progress-conditioned action memory, denoted $\mathcal{A}_{att}$, which records actions executed while the progress remains unchanged. If the predicted progress at step $n+1$ satisfies $p_{n+1} = p_n$, the executed action $a_n$ is appended to $\mathcal{A}_{att}$. This deterministic update prevents redundant retries and enables efficient exploration within a progress stage.

Some observations contain information critical for future decisions, such as intermediate results and permissible actions. Retaining all past observations would reintroduce the inefficiencies of full-history conditioning, so \gls{pabu} employs a learned observation-retention policy that predicts whether newly observed information should update the retained set $\mathcal{O}_{saved}$, which is typically sparse. Simultaneously, \gls{pabu} extracts the set of available actions $\mathcal{A}_{available}$ directly from the observation text, without assuming oracle access.

Combining progress with selectively retained context, the belief state at step $n$ is represented as
\begin{align}
b_n &= \hat{\Phi}(b_{n-1}, a_{n-1}, o_{n-1}) \nonumber \\
    &= [q, p_n, \mathcal{A}_{att}, \mathcal{A}_{available}, \mathcal{O}_{saved}],
\label{eqn:belief_pabu}
\end{align}
where the update function $\hat{\Phi}$ integrates the previous belief, the last action, and the most recent observation to produce a compact, task-relevant memory summarizing the agent’s current progress and critical context. Note that such a belief state is not a probabilistic estimate over latent environment states, but a structured abstraction that supports efficient and interpretable decision making. All components are updated autoregressively, conditioning only on the previous belief, action, and observation, which avoids the inefficiencies of full-history conditioning and enables scalable inference.

\subsection{Training and Inference Pipelines} \label{sec:alg}
We describe how the belief update and action policy are learned from data and used for inference. The goal is to equip an \gls{llm} to (i) update the belief state 
with progress and selectively retained context, and (ii) generate the progress-consistent action conditioned solely on this belief state.

\paragraph{Trajectory Augmentation.} 
We assume access to a set of successful trajectories $\mathcal{D}$, each of the form $h_n = (q, a_1, o_1, \ldots, a_n, o_n)$ with terminal reward $1$ at step $k$. Trajectories with explicit progress, subgoal, or plan annotations can use these labels directly; otherwise, supervision signals are synthesized using \gls{llm} calls at the trajectory level with human verification.

First, a set of \emph{critical actions} is identified: actions essential for task success whose removal prevents achieving the terminal reward, that are validated against the environment and used only for supervision, not during inference. This set separates the full action set into \textit{progress-consistent} if critical and non-progress-consistent groups for later action augmentation. Next, free-form progress descriptions are synthesized for each critical action, producing a progress sequence that captures task advancement. Finally, observation elements required for future actions after the current step are identified, yielding an observation-retention signal.

After augmentation, each trajectory becomes
\begin{equation*}
h' = (q, p_1, a_1, o_1, l_1, \ldots, p_k, a_k, o_k, l_k),    
\end{equation*}
where $p_i$ denotes progress and $l_i$ indicates whether to retain the observation. Belief-action pairs $(b_i, a_i)$ are then constructed by $\hat\Phi$ using these retention and progress labels, with beliefs updated autoregressively according to Eqn. \ref{eqn:belief_pabu}.

\paragraph{Progress-Aware Learning Objective.} \label{sec:training_obj}
The \gls{llm} is fine-tuned to predict the next action and the belief update components: retention, and progress, for $\hat\Phi$. Let $\mathcal{C}$ denote the critical actions in the current trajectory. The loss is
\begin{align}
\mathcal{L}_{\text{\gls{pabu}}} = &
- \sum_{a_i \in \mathcal{C}} \log P_{\pi_\theta}(l_i, p_i, a_i \mid b_i) \nonumber \\
&- \sum_{a_i \notin \mathcal{C}} \log P_{\pi_\theta}(l_i, p_i, \tilde{a}_i \mid b_i), \label{eqn:main}
\end{align}
where $\tilde{a}_i$ is the \textit{augmented action}, which is the next progress-consistent action if $a_i$ does not advance progress. Supervising the next critical action encourages the model to complete the current progress stage and transition efficiently toward the next. Training on retention and progress labels teaches the \gls{llm} $\pi_\theta$ where $\theta$ represents tunable parameters to maintain only task-relevant context. A training procedure is detailed in Algorithm \ref{alg:pabu_augment} with further discussion comparing Eqn. \ref{eqn:main} with history-based conditioning can be found in Appendix \ref{app:formulation}.

\begin{figure}[t]
    \centering
    \includegraphics[width=0.75\linewidth]{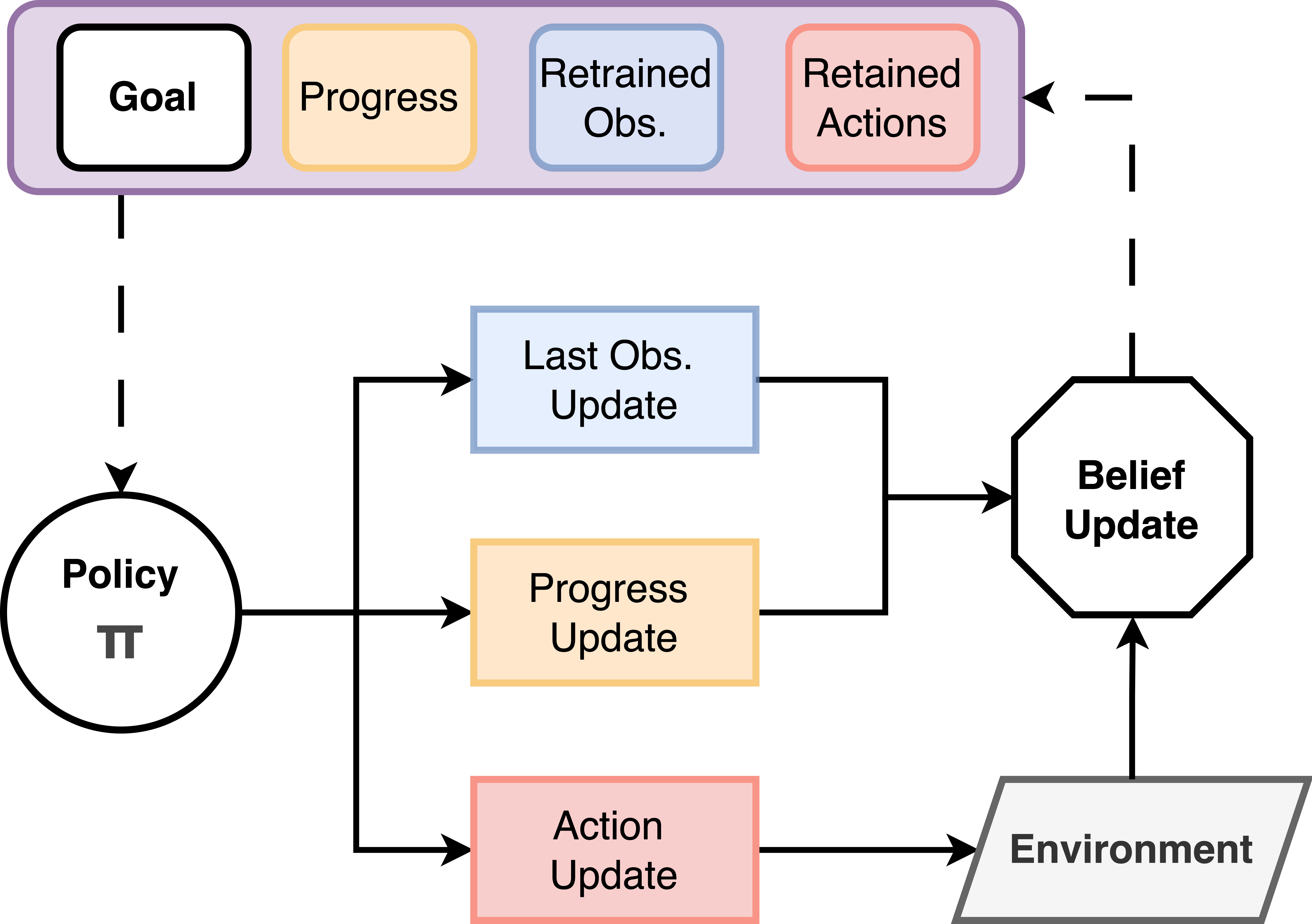}
    \caption{\textbf{\gls{pabu} Inference pipeline}, where the \gls{llm} policy generates the retention decision for the previous observation, estimates the current progress, and predicts the next action sequentially.}
    \label{fig:inference_pipeline}
\end{figure}

\paragraph{Inference Pipeline.} 
At inference, retention, progress, and action are predicted in a single \gls{llm} call per step. The order is retention $\to$ progress $\to$ action: retention depends only on past observations, which inform progress, and progress in turn guides the next action autoregressively. This loop repeats until a final-answer action is emitted or the interaction budget is exhausted, enabling efficient, full-history-free decision making (see Fig. \ref{fig:inference_pipeline}).
 
\begin{algorithm}
\caption{Progress-Aware Augmented Learning}
\label{alg:pabu_augment}
\begin{algorithmic}[1]
\STATE \textbf{Input:} Successful trajectories $\mathcal{D}$, \gls{llm} checkpoint $\pi_\theta$, progress synthesis prompt $\mathcal{P}_p$, observation-retention prompt $\mathcal{P}_o$
\STATE \textbf{Output:} Augmented dataset $\hat{\mathcal{D}}$
\STATE Initialize $\hat{\mathcal{D}} \leftarrow \emptyset$
\FOR{each trajectory $t = (q, a_1, o_1, \ldots, a_k, o_k)$ in $\mathcal{D}$}
    \STATE Identify critical actions $\mathcal{C} \subseteq \{a_1, \ldots, a_k\}$ and synthesize progress sequence $(p_1, \ldots, p_k)$ using $\mathcal{P}_p$
    \STATE Predict obs. retention $(l_1, \ldots, l_k)$ using $\mathcal{P}_o$
    \STATE Let $\mathcal{A}_{att}^{(i)} = \varnothing$, $\mathcal{O}_{saved}^{(i)} = \varnothing$, initial belief $b_0$
    \FOR{$i = 1$ to $k$}
        \STATE Select and execute action $\hat{a}_i$ from $\{a_i, \tilde{a}_i\}$, extract observation $o_i$ and retention label $l_i$
        \STATE Add training example $(b_i, \hat{a}_i, l_i)$ to $\hat{\mathcal{D}}$
        \STATE Update $\mathcal{A}_{att}^{(i+1)}$ and $\mathcal{O}_{saved}^{(i)}$ from $\hat{a}_i$
        \STATE Update $\mathcal{A}_{available}^{(i+1)}$ from $o_i$
        \STATE Set $b_{i+1} = \hat\Phi \left[b_{i}, (\mathcal{A}_{att}^{(i+1)}, \mathcal{A}_{available}^{(i+1)}) , \mathcal{O}_{saved}^{(i)} \right]$
    \ENDFOR
\ENDFOR
\STATE Update the policy checkpoint: $\pi_{\theta'} \gets \pi_{\theta}$ using $\hat{\mathcal{D}}$
\RETURN $\pi_{\theta'}$, $\hat{\mathcal{D}}$
\end{algorithmic}
\end{algorithm}

\begin{table*}[t]
\centering
\small
\caption{\textbf{Evaluating task completion and efficiency on the AgentGym suite.}
For each method, we report the task completion rate ($\uparrow$ succ \%, \textit{higher is better}) and the number of \blue{interaction steps} ($\downarrow$ step \#, \textit{lower is better}). Aggregated performance across eight tasks is summarized in the $\sum$ column, and the best-performing model is highlighted in \textbf{bold}.}
\label{tab:main_results}
\setlength{\tabcolsep}{5pt}
\begin{tabular}{l|l|rrrrrrrr|r}
\toprule
\textbf{Model} & \textbf{Metric}& \textbf{ALF} & \textbf{TC} & \textbf{Sci} & \textbf{Baby} & \textbf{MZ} & \textbf{WD} & \textbf{WT} & \textbf{MV} & \multicolumn{1}{c}{$ \sum $}   \\

\midrule
\multicolumn{11}{c}{\textbf{API-based Models \& Agents}} \\
\midrule

\multirow{2}{*}{DeepSeek-Chat}
& $\uparrow$ succ \%  & 51.0 & 23.0 & 16.8 & 45.7 & 4.0 & 24.0 & 70.0 & 70.0 & 34.5  \\
& $\downarrow$ step \# & \blue{20.4} & \blue{15.1} & \blue{20.7} & \blue{11.7} & \blue{14.5} & \blue{5.2} & \blue{6.1} & \blue{5.9} & \blue{16.9} \\

\multirow{2}{*}{Claude-3-Sonnet}
& $\uparrow$ succ \%  & 13.0 & 38.0 & 2.8 & 79.3 & 0.0 & 36.0 & 65.0 & 80.0 & 26.3 \\
& $\downarrow$ step \# & \blue{27.9} & \blue{14.6} & \blue{28.7} & \blue{6.6} & \blue{15.0} & \blue{5.2} & \blue{6.9} & \blue{5.1} & \blue{20.8} \\

\multirow{2}{*}{GPT-4-Turbo}
& $\uparrow$ succ \%  & 67.5 & 77.0 & 14.4 & 72.9 & 68.0 & 88.0 & 80.0 & 95.0 & 55.9   \\
& $\downarrow$ step \# & \blue{18.3} & \blue{9.9} & \blue{18.1} & \blue{9.1} & \blue{9.0} & \blue{4.0} & \blue{6.0} & \blue{4.5} & \blue{14.2} \\

\midrule
\multicolumn{11}{c}{\textbf{General-purpose Models \& Agents}} \\
\midrule

\multirow{2}{*}{Llama2-Chat-7B}
& $\uparrow$ succ \%  & 2.0 & 0.0 & 0.8 & 0.2 & 0.0 & 0.0 & 0.0 & 0.0 & 0.9  \\
& $\downarrow$ step \# & \blue{22.6} & \blue{14.5} & \blue{27.5} & \blue{9.5} & \blue{15.0} & \blue{6.0} & \blue{9.9} & \blue{12.0} & \blue{19.5} \\

\multirow{2}{*}{Llama2-Chat-13B}
& $\uparrow$ succ \%  & 3.5 & 0.0 & 0.8 & 0.1 & 0.0 & 0.0 & 0.0 & 0.0 & 1.3 \\
& $\downarrow$ step \# & \blue{19.6} & \blue{16.5} & \blue{21.3} & \blue{10.9} & \blue{13.4} & \blue{6.0} & \blue{10.0} & \blue{12.0} & \blue{17.3}\\

\multirow{2}{*}{AgentLM-7B}
& $\uparrow$ succ \%  & 71.0 & 4.0 & 1.6 & 0.5 & 12.0 & 4.0 & 0.0 & 5.0 & 22.7  \\
& $\downarrow$ step \# & \blue{17.7} & \blue{19.4} & \blue{28.5} & \blue{7.5} & \blue{13.9} & \blue{2.0} & \blue{8.3} & \blue{11.7} & \blue{18.6} \\

\multirow{2}{*}{AgentLM-13B}
& $\uparrow$ succ \%  & 73.0 & 0.0 & 2.8 & 0.5 & 8.0 & 0.0 & 10.0 & 5.0 & 23.1\\
& $\downarrow$ step \# & \blue{17.8}  & \blue{19.4}  & \blue{28.5}  & \blue{7.6}  & \blue{13.9}  & \blue{6.0}  & \blue{6.6}  & \blue{10.7}  & \blue{18.7}  \\

\multirow{2}{*}{AgentLM-70B}
& $\uparrow$ succ \%  & 67.0 & 4.0 & 10.7 & 0.7 & 8.0 & 4.0 & 0.0 & 0.0 & 24.0  \\
& $\downarrow$ step \# & \blue{18.5}  & \blue{18.8}  & \blue{28.2}  & \blue{6.3}  & \blue{13.9}  & \blue{5.2}  & \blue{6.6}  & \blue{11.6}  & \blue{18.6}  \\

\midrule
\multicolumn{11}{c}{\textbf{Benchmark-tuned Models \& Agents}} \\
\midrule
\multirow{2}{*}{AgentEvol-7B}
  & $\uparrow$ succ \%  & {\textbf{88.0}} & {64.0} & {38.0} & {82.7} & {12.0} & {12.0} & {25.0} & {60.0}  & 60.8\\
   & $\downarrow$ step \# & \blue{14.0} & \blue{{11.8}} & \blue{18.9} & \blue{{4.3}} & \blue{13.8} & \blue{{5.7}} & \blue{5.9} & \blue{\textbf{3.2}}  & \blue{{13.0}} \\

\multirow{2}{*}{ATL\scalebox{0.7}{A}S-8B}
 & $\uparrow$ succ \%   & {84.5} & {{72.0}} & {42.0} & {{81.0}} & {{48.0}} & {20.0} & {\textbf{60.0}} & {\textbf{90.0}} & {65.4} \\
  & $\downarrow$ step \#  & \blue{-} & \blue{-} & \blue{-} & \blue{-} & \blue{-} & \blue{-} & \blue{-} & \blue{-}& \blue{-} \\

\multirow{2}{*}{\textbf{\gls{pabu}-8B(ours)}}
 & $\uparrow$ succ \%   & {86.5} & \textbf{73.0} & {\textbf{85.5}} & \textbf{94.4} & \textbf{68.0} & {\textbf{40.0}} & {35.0} & {75.0} & \textbf{81.0} \\
 & $\downarrow$ step \#   & \blue{\textbf{11.0}} & \blue{\textbf{8.9}} & \blue{\textbf{12.5}} & \blue{\textbf{3.8}} & \blue{\textbf{9.7}} & \blue{\textbf{5.4}} & \blue{\textbf{5.1}} & \blue{4.0} & \blue{\textbf{9.5}} \\

\bottomrule
\end{tabular}
\end{table*}

\section{Numerical Study}
We conduct our experiments in the AgentGym Suites \citep{xi2024agentgym}, which provide a diverse set of environments and tasks designed for broad, real-time, unified-format, and concurrent \gls{llm} agent evaluation. AgentGym enables systematic benchmarking of agentic reasoning and interaction capabilities across heterogeneous tasks. Due to page limits, we highlight the key findings here; additional implementation details are provided in Appendix \ref{app:experiment}.

\subsection{Baselines and Metrics}
\label{sec::metrics}
We compare our proposed \gls{llm} agent against three categories of baselines: (1) proprietary API-based models and agents, (2) open-source, reproducible models and agents with diverse architectural designs, and (3) \gls{llm} agents and models explicitly optimized for AgentGym environments. A comprehensive description of all baselines is provided in Appendix \ref{app:baseline}.

For evaluation, we primarily report the task success rate (succ\%) and completion efficiency, measured as the expected number of interaction steps required to complete a task (step\#). The latter is computed by averaging episode lengths over both successful and failed trajectories, providing a holistic measure of agent efficiency. Additionally, in ablation studies, we analyze input/output token counts and inference wall-clock time to better understand performance trade-offs. Formal definitions and computation details for all metrics are provided in Appendix \ref{app:metric}.

\begin{figure*}[t]
    \centering
    \includegraphics[width=0.75\linewidth]{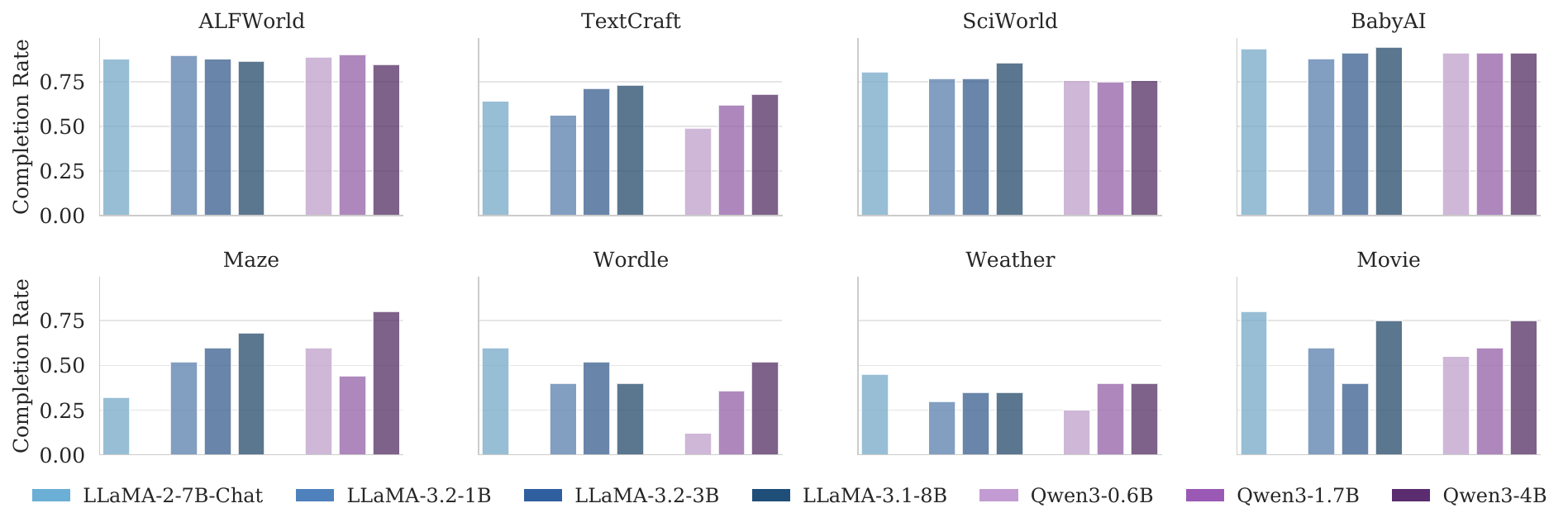}
    \caption{\textbf{Performance comparison across different backbone models.}
Smaller models can learn embodied environments effectively given sufficient training samples, while larger models generally perform better in environments that require deeper language understanding.}
    \label{fig:abla_size}
\end{figure*}

\begin{figure*}[t]
    \centering
    \includegraphics[width=0.75\linewidth]{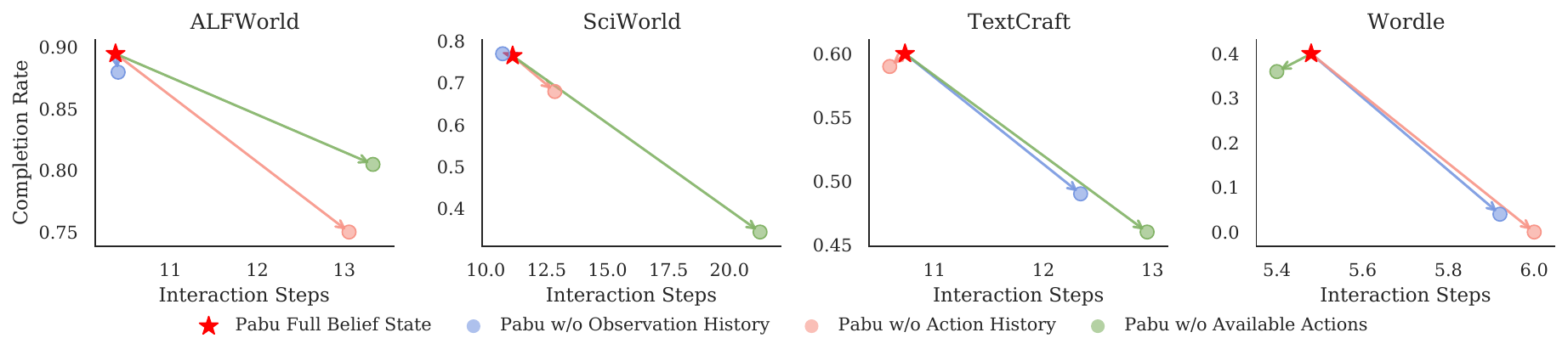}
    \caption{\textbf{Effect of context ablations on efficiency and performance.} Each point shows the interaction steps and task completion rate after removing a context component. Retaining observation history is critical for tasks that require long-term referencing of multiple observations, action history is most important in environments where similar observations may require diverse responses, while removing available actions disproportionately degrades performance in environments with evolving action spaces. These results illustrate how task structure shapes optimal context-selection strategies.}
    \label{fig:aba_component}
\end{figure*}

\subsection{Main Results}
\label{main::results}
Table~\ref{tab:main_results} summarizes the performance of different models on the AgentGym environments. Results for open-source models are taken directly from their original publications without reproduction or reimplementation. Unreported metrics are denoted by -.

Overall, the results demonstrate that models fine-tuned on the AgentGym Suites substantially outperform their corresponding pre-trained checkpoints and consistently surpass general-purpose models by a clear margin, highlighting the importance of task-aligned agentic training for improved controllability and performance. Among proprietary models evaluated on AgentGym, GPT-4-turbo outperforms DeepSeek-v3 and Claude-3, achieving the highest task success rate (55.9\%) while requiring the fewest interaction steps (14.2). For open-weight general-purpose models, the AgentLM family exhibits the strongest agentic capabilities; in particular, the 70B variant attains a 24.0\% success rate with an average of 18.6 interaction steps.

We next consider models explicitly trained on the AgentGym Suites, all of which are trained on the same trajectory collection, AgentTraj-L, but with different training procedures. AgentEvol, which combines supervised fine-tuning and reinforcement learning, achieves a 60.8\% success rate with an average of 13.0 interaction steps. ATL\scalebox{0.7}{A}S, which selectively trains on a subset of actions (30\%), further improves performance to a 65.4\% aggregated success rate, although the corresponding number of interaction steps is not reported. Finally, our proposed \gls{pabu} achieves the best overall performance, attaining an 81.0\% task success rate with only 9.5 interaction steps on average, corresponding to relative improvements of 23.9\% in success rate and 26.9\% in efficiency over the current state of the art.

\subsection{Backbone Model Ablation Study}

We first examine whether \gls{pabu} generalizes across backbone models from different families and at varying scales. Specifically, we apply \gls{pabu} to \gls{llm}s from the \textsc{Llama} and \textsc{Qwen} families (see Appendix \ref{app:baseline} for details). Notably, we also evaluate Llama-2-7B-Chat, which serves as the backbone model in \citet{xi2024agentgym}, to control for potential baseline performance differences. All models are trained with \gls{pabu} and evaluated under the same experimental settings as in the main experiments. Figure~\ref{fig:abla_size} shows side-by-side performance comparisons across benchmarks, and the full set of metrics is provided in Appendix \ref{app:size_abla_perf}.

Overall, \gls{pabu} consistently improves agentic performance across backbone families and model sizes, demonstrating strong robustness to architectural and scale variations. Notably, the older Llama-2-7B-Chat checkpoint also achieves a comparable task completion rate of 78.2\%. While larger models 
yield better task performance, they incur substantially higher computational overhead. To strike a balance between efficiency and performance, we adopt Llama-3.2-1B as the backbone model for subsequent experiments.

\begin{table*}
\centering
\caption{\textbf{Attribution Analysis across Training Procedures.} Progress-based action filtering (ORM$\rightarrow$PRM) and augmentation (PRM$\rightarrow$progress-based action augmentation) consistently improve performance under history-based belief representations. Incorporating \gls{pabu} further refines belief design and yields additional gains beyond next decision filtering and augmentation alone.}
\scalebox{0.8}{%
\begin{tabular}{lrcccccrr}
\hline
\multirow{2}{*}{\textbf{Training Objective}} 
& \multicolumn{3}{c}{\textbf{Training Supervision}} 
& \multicolumn{3}{c}{\textbf{Test Performance}} 
& \multicolumn{2}{c}{\textbf{Token Cost}} \\
& samples & belief rep. & action 
& succ \% & step \# & time 
& input & output \\  
\hline
Outcome Reward Model 
& 28,162 & History & Original 
& 48.5 & 14.5 & 57.10 
& 1,672,162 & 22,219 \\
Progress Reward Model
& 12,963 & History & Filtered 
& 80.5 & 11.9 & 33.87 
& 1,389,159 & 18,495 \\
Progress Augmentation          
& 28,162 & History & Augmented 
& 84.5 & 8.2 & 14.97 
& 1,015,571 & \textbf{15,042} \\
\gls{pabu} w/ Augmentation   
& 28,162 & \gls{pabu} & Augmented 
& \textbf{90.0} & \textbf{8.1} & \textbf{12.51} 
& \textbf{502,426} & 25,869 \\
\hline
\end{tabular}}
\label{abla::formulation}
\end{table*}

\subsection{Attribution on Context Components} \label{sec:contextcomp}

In this experiment, we quantify the empirical contributions of historical actions and observations to gain insights into effective context management strategies. Specifically, we train separate models in which available actions, historical actions, or observations are masked throughout training and evaluation, and attribute performance differences to the removed components.

Results are reported in Fig. \ref{fig:aba_component} and Table \ref{app::abla_compoent_tab}. We find that sensitivity to context components varies substantially across environments, reflecting differences in task structure. In the AgentGym suite, ALFWorld and ScienceWorld do not necessarily need prior observations (e.g., inventory information), but both action history and available actions are important due to environment-specific action constraints. In Wordle, the action space is fixed, enabling generalization from successful trajectories; however, prior guesses (actions) and feedback (observations) are essential for correct reasoning. In TextCraft, actions are largely deterministic (e.g., following recipes), making action history less critical, whereas retaining observation history and available actions significantly improves performance. 

Overall, these results suggest that (1) observation history is most important for tasks requiring cross-referencing prior responses, (2) action history is crucial in environments where identical queries admit different correct actions, and (3) explicit modeling of the available action space is most beneficial when permissible actions vary across task instances, highlighting how task structure shapes optimal context-selection strategies.

\subsection{Attribution on Training Procedures} \label{exp:training_obj}
To empirically disentangle the contributions of the training procedures introduced in Section \ref{sec:training_obj}, we conduct an ablation study on ALFWorld with progressively stronger supervision and decision augmentation. The baseline model uses the full interaction history as the belief state and is trained on decision actions from trajectories with outcome reward equal to $1$. Following \citet{chen2025atlasagenttuninglearning}, we then incorporate the progress-advancement signal to filter decisions and their associated history, approximating a progress reward model. Next, we augment the full-history belief state with progress-advancing actions to further enrich training supervision. Finally, we evaluate \gls{pabu}, which integrates progress-aware belief modeling with action augmentation.

Across these incremental settings shown in Table \ref{abla::formulation}, we observe consistent performance improvements that progress signals are effective for both action filtering and belief state augmentation. In particular, \gls{pabu} achieves higher task success rates while requiring fewer token consumptions and interaction steps, demonstrating that jointly modeling progress-aware beliefs and action augmentation, as in \gls{pabu}, yields complementary benefits beyond isolated techniques.

\section{Limitations and Future Directions}
We highlight aspects of \gls{pabu} for further study and defer these extensions to future work. To facilitate follow-up research, we will open-source our model checkpoints, environments, and training recipes upon publication.

\textbf{Data Coverage and Exploration.} \gls{pabu} operates as an offline optimization method that enables \gls{llm} agents to manage their own beliefs, and its performance can benefit from diverse and representative seed trajectories. As the current framework does not incorporate native online exploration during training, adaptivity under rare failure cases or previously unseen observations may be limited. Integrating \gls{pabu} with complementary online or iterative data collection strategies, such as self-play or environment-driven exploration, could further improve robustness and generalization.

\textbf{State Abstraction and Representation.} The \gls{pabu} framework aims to learn improved representations or abstractions of partially observed environment states. In this work, we instantiate this objective through synthesized progress, which provides a practical and effective realization but may not fully capture all aspects of optimal abstraction. Exploring alternative abstraction mechanisms, such as hierarchical representations or learned latent-state models, is a natural direction for future research. 

\section{Conclusion}

We presented \gls{pabu}, a general framework for efficient \gls{llm} agents operating in \gls{pomdp} settings. By explicitly modeling task progress through progress-aware belief updates, \gls{pabu} enables agents to retain only task-relevant historical information while avoiding long-range summarization, thereby improving both decision accuracy and computational efficiency by prioritizing progress-consistent actions on all explored states. Experiments on \textit{AgentGym} demonstrate substantial gains in task success rates and reduced interaction steps compared to strong baselines, highlighting the effectiveness of progress-aware history abstraction.

Beyond overall performance improvements, our ablation studies show that the benefits of \gls{pabu} generalize across different backbone architectures and model scales, with individual components contributing differently depending on task characteristics. These results suggest that explicit reasoning about task progress constitutes a broadly applicable design principle for building scalable and efficient LLM-based agents.

\clearpage

\appendix
\section*{Acknowledgment}
The authors gratefully acknowledge Jun Wang for valuable discussions on the preliminary problem formulation and Shengjie Liu for suggestions on the implementation. The authors will also appreciate the constructive feedback from the anonymous ICML reviewers. In addition, we thank Freepik (\url{www.freepik.com}) for the chatbot icon designs.

\bibliography{icml_cite,application}
\bibliographystyle{icml2026/icml2026}

\onecolumn

\section{Comparing Pabu and History-based Formulation} \label{app:formulation}
Efficient reasoning in LLM agents requires focusing on actions that meaningfully advance task progress while avoiding redundant or irrelevant steps. We achieve this by identifying critical actions, augmenting trajectories with progress information, and training the model to prioritize \textbf{progress-consistent actions}. This section describes how full-history imitation is extended into a trajectory-level, progress-aware refinement framework with numerical support from an ablation study in Section \ref{exp:training_obj}.

\paragraph{From Full Trajectory Imitation to Progress-Consistent Actions.}  
A common approach is to imitate every action in successful trajectories:
\begin{equation}
\mathcal{L}_{\text{ORM}} = - \sum_{i\le n} \log P_{\pi_\theta}(a_i \mid h_{i-1}).
\end{equation}
Alternatively, one can focus only on actions leading to \textbf{critical states}:
\begin{equation}
\mathcal{L}_{\text{PRM}} = - \sum_{a_i \in \mathcal{C}} \log P_{\pi_\theta}(a_i \mid h_{i-1}).
\end{equation}
Unlike PRM derived from MCTS, this history-based setup may include prior actions with zero progress reward before the next progress-consistent action. Learning from all actions risks replicating unnecessary steps, whereas learning only from critical actions can omit knowledge of intermediate transitions needed to reach those states. \textbf{Progress-aware refinement} addresses both issues by guiding the model to skip avoidable steps while preserving essential context.

\paragraph{Trajectory-Level Action Refinement.}  
Consider a trajectory segment from the underlying MDP: $h_2 = (q, a_1, s_1, a_2, s_2)$ with cumulative cost 
$C_1 = \mathcal{C}(a_1, h_0) + \mathcal{C}(a_2, h_1)$, 
where $h_0 = (q)$ and $h_1 = (q, a_1, s_1)$. Suppose a \textbf{refined action} $a_{[1]}$ exists such that:
\begin{equation*}
e(h_0, a_{[1]}) = s_2 \quad \text{and} \quad \mathcal{C}(a_{[1]}, h_0) < C_1.
\end{equation*}
Encouraging the agent to take $a_{[1]}$ bypasses the intermediate state $s_1$, producing a \textbf{synthesized trajectory} with lower cumulative cost. This captures the essence of progress-aware reasoning: the model skips unnecessary steps while still reaching subsequent critical states efficiently.

\paragraph{Learning Refined, Progress-Consistent Actions.}  
Given a trajectory $h_n = (q, a_1, o_1, \dots, a_n, o_n)$ and a set of critical actions $\mathcal{C}$, we define the \textbf{Progress Augmentation loss}:
\begin{align}
\begin{split}
\mathcal{L}_{\text{PA}} = & - \sum_{a_{i} \in \mathcal{C}} \log P_{\pi_\theta}(a_i \mid h_{i-1}) \\
& - \sum_{a_{i} \notin \mathcal{C}} \log P_{\pi_\theta}(a_{[i]} \mid h_{i-1}),
\end{split}
\end{align}
where $a_{[i]}$ is the \textbf{next progress-consistent action} toward overall task completion. This encourages the policy to prioritize actions that advance progress while skipping avoidable intermediate steps.

\paragraph{\gls{pabu} Loss for Progress-Aware Learning.}  
In \gls{pabu}, the model is fine-tuned to predict the next action along with belief update components, \textbf{retention} and \textbf{progress}, for the current belief $b_i$. Let $\mathcal{C}$ denote the critical actions in a trajectory. The loss is:
\begin{align}
\mathcal{L}_{\text{\gls{pabu}}} = & - \sum_{a_i \in \mathcal{C}} \log P_{\pi_\theta}(l_i, p_i, a_i \mid b_i) \nonumber \\
& - \sum_{a_i \notin \mathcal{C}} \log P_{\pi_\theta}(l_i, p_i, \tilde{a}_i \mid b_i),
\end{align}
where $\tilde{a}_i$ is the \textit{augmented action}, i.e., the next progress-consistent action if $a_i$ does not advance progress. Supervising critical actions ensures the model completes the current progress stage efficiently and transitions toward the next. Training on retention and progress labels enables the model to maintain only task-relevant context, improving both efficiency and robustness.

\section{Experiment Reproduction Statement} \label{app:experiment}
All experiments are conducted by training on servers equipped with NVIDIA H100 (or GH200) GPUs and evaluating on a local machine with an Intel i5-9600K CPU, 32GB RAM, and an RTX 3090 GPU, running Ubuntu 22.04 LTS and Python 3.11. Detailed environment setup instructions, dependency versions, and reproducibility scripts are provided in the accompanying support materials.

\subsection{Experiment Environments} \label{app:process_summary}
We adopt all multi-step environments from AgentGym that provide reproducible training trajectories (excluding To-Do here as each rerun generates trajectories with different ids) and avoid potential conflicts of interest (e.g., WebShop \citep{yao2023webshopscalablerealworldweb}). The following environment descriptions follow the definitions in \citet{xi2024agentgym}\footnote{\url{https://github.com/WooooDyy/AgentGym/blob/main/LICENSE}}. As these environments and AgentTraj-L\footnote{\url{https://huggingface.co/datasets/AgentGym/AgentTraj-L/tree/main}} are prior works, we briefly summarize their setups and focus on our critical step identification and progress synthesis procedures.

\paragraph{MAZE (MZ). \citep{abdulhai2023lmrlgymbenchmarksmultiturn}}
MAZE is a grid-based word game in which agents navigate toward a goal using four directional actions, incurring a reward of $-1$ per step until success. We use AgentTraj-L trajectories filtered to exclude trivial starts at the goal, with a maximum episode length of 15. Progress is heuristically synthesized from the agent’s Manhattan distance to the goal, providing a continuous notion of advancement. As the environment is fully observable and memory-free, no observation retention mechanism is applied. Critical steps are identified as moves that strictly reduce the distance to the goal or resolve navigation dead-ends.\footnote{\url{https://github.com/abdulhaim/LMRL-Gym/blob/main/LICENSE}}

\paragraph{Wordle (WD). \citep{abdulhai2023lmrlgymbenchmarksmultiturn}}
Wordle evaluates letter-level reasoning through iterative word guesses, with a reward of $-1$ per attempt until the correct word is found. AgentTraj-L trajectories are directly reused. Since Wordle does not admit a natural scalar notion of intermediate progress, no explicit progress estimation is applied. Observation retention instead focuses on translating structured feedback (e.g., ``g y b b y'') into persistent constraints over letter inclusion, exclusion, and positional validity. Critical steps correspond to guesses that introduce new constraints or eliminate large portions of the candidate space.\footnote{\url{https://github.com/abdulhaim/LMRL-Gym/blob/main/LICENSE}}

\paragraph{ALFWorld (ALF). \citep{shridhar2020alfworld}}
ALFWorld is a household task environment requiring exploration and object manipulation through text-based interactions. We use AgentTraj-L trajectories collected from state-of-the-art models and human demonstrations. For each trajectory, both critical actions and progress summaries are synthesized using a unified LLM prompt (Llama-3.3-70B). Critical steps are identified as irreversible or goal-enabling actions (e.g., acquiring a required object or unlocking a location), while progress synthesis abstracts low-level action sequences into task-relevant subgoals, yielding a compact semantic trajectory.\footnote{\url{https://github.com/alfworld/alfworld/blob/master/LICENSE}}

\paragraph{SciWorld (Sci). \citep{wang2022scienceworldagentsmarter5th}}
ScienceWorld targets elementary-level scientific reasoning across diverse experimental tasks. Due to the availability of golden paths and the difficulty of achieving strong performance with existing agents, we generate trajectories by prompting GPT-4-Turbo to produce reasoning traces aligned with these golden paths. Progress is heuristically synthesized from task-specific reward signals and experiment milestones (e.g., successful measurement or setup completion). Critical actions are first annotated by humans and then filtered using an LLM-based consistency check, focusing on actions unlocking subsequent steps.\footnote{\url{https://github.com/allenai/ScienceWorld/blob/main/LICENSE}}

\paragraph{BabyAI (Baby). \citep{chevalierboisvert2019babyaiplatformstudysample}}
BabyAI is a partially observable grid world with instruction-following tasks. We use the AgentBoard textual abstraction, which converts visual observations into language and exposes high-level actions. AgentTraj-L trajectories are reused with a maximum episode length of 20. Critical steps are identified heuristically as actions that complete sub-instructions (e.g., picking up a specified object or opening a required door). Progress is synthesized using a scene-based abstraction that tracks instruction satisfaction and environment state transitions rather than raw movement actions.\footnote{\url{https://github.com/mila-iqia/babyai/blob/master/LICENSE}}

\paragraph{TextCraft (TC). \citep{prasad2024adaptasneededdecompositionplanning}}
TextCraft is a text-only crafting environment derived from Minecraft recipes, where tasks require compositional planning over up to four crafting steps. AgentTraj-L trajectories are used with a maximum of 20 rounds. Critical actions are defined as successful item acquisitions and crafting operations that advance the recipe tree. Progress synthesis abstracts the current inventory and crafted items into the next required recipe step, yielding a forward-looking representation of task completion status rather than a retrospective action log.\footnote{\url{https://github.com/archiki/ADaPT/blob/main/LICENSE}}

\paragraph{Weather (WT). \citep{ma2024agentboardanalyticalevaluationboard}}
The Weather environment enables agents to query structured weather information via tool calls. Trajectories from AgentTraj-L are reused, with success evaluated by exact answer matching and a maximum of 10 rounds. Critical steps are identified as tool invocations that return valid (non-error) responses, contributing necessary information for the final answer. Observation retention and progress synthesis are determined using an LLM prompt (gpt-oss-120B), which filters redundant queries and summarizes acquired information into a compact knowledge state.\footnote{\url{https://github.com/hkust-nlp/AgentBoard}. The codebase is licensed under an Apache-2.0 License and the dataset is licensed under a GNU General Public License, version 2.}

\paragraph{Movie (MV). \citep{ma2024agentboardanalyticalevaluationboard}}
The Movie environment allows agents to query cinematic metadata via a movie database API. We reuse AgentTraj-L trajectories annotated by GPT-4-Turbo, with a maximum episode length of 12. Similar to Weather, critical actions are tool calls yielding valid responses that introduce new entities or relations relevant to the query. Progress synthesis is performed via an LLM prompt (gpt-oss-120B) that aggregates retrieved facts and tracks remaining informational gaps toward answering the question.\footnote{\url{https://github.com/hkust-nlp/AgentBoard}}

\subsection{Processed Dataset Statistics}
Table \ref{tab:stat} summarizes the number of trajectories and decision steps for each environment in the full dataset used in this work.
For the main experiment and the first two ablation studies, the training datasets are either directly taken from or transformed versions of this full dataset.
In the final ablation study, we restrict training to the ALFWorld dataset only.
This choice is motivated by the fact that ALFWorld is not sensitive to the observation retention mechanism, allowing us to isolate and evaluate the effect of different training objectives without confounding factors.

\begin{table}[h]
\caption{Number of trajectories and decision counts per dataset}
\label{tab:stat}
\centering
\begin{tabular}{lcccccccc|c}
\hline
 & \textbf{ALF} & \textbf{TC} & \textbf{Sci} & \textbf{Baby} & \textbf{MZ} & \textbf{WD} & \textbf{WT} & \textbf{MV} & $\sum$ \\
\hline
\textbf{\# Trajectories} 
& 2164 & 374 & 1986 & 761 & 10 & 955 & 311 & 215 & 6776 \\

\textbf{\# Decisions} 
& 39775 & 5084 & 63855 & 7022 & 528 & 7134 & 3312 & 1682 & 128392 \\
\hline
\end{tabular}
\end{table}

\subsection{Baseline Models} \label{app:baseline}
For the main experiments, a range of closed-source models is evaluated, including GPT-3.5-Turbo, GPT-4-Turbo, Claude-3, and DeepSeek-Chat \citep{deepseekai2025deepseekv3technicalreport}.
We also benchmark open-source models such as Llama-2 \citep{touvron2023llama2openfoundation}, as well as agents trained on expert trajectories, namely AgentLM \citep{zeng2023agenttuningenablinggeneralizedagent}.

For fine-tuned backbones, we use Llama-3.1-8B \citep{grattafiori2024llama3herdmodels} as the primary model in the main experiments.
For the model size ablation study, we evaluate Llama-3.2 models at 1B and 3B scales, Llama-2-7B-Instruct, and Qwen3 models \citep{yang2025qwen3technicalreport} at 0.7B, 1.6B, and 4B scales.
For the component and learning objective ablation, we adopt Llama-3.2-1B as the backbone.

\subsection{Evaluation Metrics} \label{app:metric}

For the main experiment, we report the task completion rate (succ.\%) and the number of interaction steps (Steps). The task completion rate is defined as the proportion of tasks that are successfully completed with an environment return of 1 within the predefined step limit. The number of interaction steps counts the number of interaction rounds required to complete a task, with the step limit serving as an upper bound for unsuccessful cases. Aggregated performance is computed using the macro average across different settings for each environment.

For the backbone model ablation and context ablation studies, we use the same evaluation metrics as those in the main experiment.

For the training procedure ablation, we additionally report the number of training samples (Samples), the type of belief representation (Belief Rep.), and the action acquisition method used to extract actions from the original trajectories (Action). We also report the task completion rate together with the wall-clock time required to complete the evaluation. Finally, we report the total number of input and output tokens, counted within the relevant XML tags, aggregated over all evaluated cases.

\subsection{Prompt Examples}
Figure \ref{fig:alfworld_prompt} illustrates the prompt used in ALFWorld for extracting critical actions and synthesizing progress from example trajectories, showing how target goals are decomposed into actionable steps with progresses (sub-goals).

\begin{figure}
\centering
\begin{tcolorbox}[colback=purple!5!white,colframe=purple!35!black, title=ALFWorld Critical Action Selection and Progress Synthesis Prompt]
\begin{lstlisting}
Given example key steps with their sub-goals for finishing an example goal, identify the key steps in a trajectory with the format of "action -> observation" for accomplishing a target goal. Directly reply in bullets without comments.

Example goal:
find two book and put them in bed.

Example plan with sub-goals:
- go to desk 1 -> find the book (1 of 2)
- take book 3 from desk 1 -> take the book (1 of 2)
- go to bed 1 -> go to the bed (1 of 2)
- put book 3 in/on bed 1 -> put the book in the bed (1 of 2)
- go to drawer 1 -> find the book (2 of 2)
- open drawer 1 -> take the book (2 of 2)
- take book 2 from drawer 1 -> go to the bed (2 of 2)
- go to bed 1 -> put the book in the bed (2 of 2)
- put book 2 in/on bed 1 -> put the book in the bed (2 of 2)

Target goal:
{g}

Trajectory:
{t}

Key steps with sub-goals:
\end{lstlisting}
\end{tcolorbox}
\caption{Critical Step Identification and Progress Synthesis for ALFWorld. Here \textbf{g} and \textbf{t} serve as two placeholders, which is formatted with the actual goal and the trajectory during API calling.}
\label{fig:alfworld_prompt}
\end{figure}

\subsection{LLM Input-Output Format and Context Management Logic}
As shown in Figure \ref{alfworld_io}, the agent maintains a structured \gls{pabu} belief state including goal, progress, last observation, and available actions, and then outputs updates to its observations, progress, and next action.

\begin{figure}
\centering
\begin{tcolorbox}[colback=blue!6!white,colframe=blue!75!black, title=\gls{pabu} Environment Input Format]
\begin{lstlisting}
<goal>
clean some knife and put it in diningtable.
</goal>

<progress>
find the knife
</progress>

<last_observation>
You open the cabinet 3. The cabinet 3 is open. In it, you see a dishsponge 2.
</last_observation>

<saved_observation>
{}
</saved_observation>

<available_action>
close cabinet 3,examine cabinet 3,go to cabinet 1,go to cabinet 10,
go to cabinet 2,go to cabinet 4,go to cabinet 5,go to cabinet 6,
go to cabinet 7,go to cabinet 8,go to cabinet 9,go to coffeemachine 1,
go to countertop 1,go to diningtable 1,go to drawer 1,go to drawer 2,
go to fridge 1,go to garbagecan 1,go to microwave 1,go to sinkbasin 1,
go to stoveburner 1,go to stoveburner 2,go to stoveburner 3,
go to stoveburner 4,go to toaster 1,inventory,look,
take dishsponge 2 from cabinet 3
</available_action>

<attempted_action>
go to drawer 1,go to cabinet 1,open cabinet 1,
go to cabinet 2,open cabinet 2,go to cabinet 3,open cabinet 3
</attempted_action>

<observation_update>
\end{lstlisting}
\end{tcolorbox}
\begin{tcolorbox}[colback=blue!5!white,colframe=blue!75!black, title=\gls{pabu} Environment Output Format]
\begin{lstlisting}
None
</observation_update>

<progress_update>
find the knife
</progress_update>

<action_update>
go to diningtable 1
</action_update>
\end{lstlisting}
\end{tcolorbox}
\caption{\textbf{Agent Input-output format.} Everything before the \textit{observation\_update} xml tag is the proposed \gls{pabu} belief state, and after it are the expected model responses.}
\label{alfworld_io}
\end{figure}

\subsection{Model Size Ablation Performance} \label{app:size_abla_perf}
Table \ref{tab:size_abla_detail} provides a detailed breakdown of final task completion rates and action counts for each backbone model across all evaluated environments. The results show that \gls{pabu} maintains strong performance across different model families (\textsc{Llama} and \textsc{Qwen}) and scales, with some variation depending on environment complexity. Smaller models like Llama-3.2-1B achieve competitive completion rates with lower action counts, highlighting their efficiency, while larger models generally show modest improvements at the cost of increased computation. These findings complement the summary in the main text and support the choice of Llama-3.2-1B as the backbone for subsequent experiments.

\begin{table}
\centering
\caption{Model Scale × Environment Results}
\label{tab:size_abla_detail}
\setlength{\tabcolsep}{5pt}
\begin{tabular}{l|l|rrrrrrrr}
\toprule
\textbf{Model} & \textbf{Metric} & \textbf{ALF} & \textbf{TC} & \textbf{Sci} & \textbf{Baby} & \textbf{MZ} & \textbf{WD} & \textbf{WT} & \textbf{MV} \\
\midrule
\multirow{2}{*}{LLaMA-3.2-1B}
 & Final label   & 89.5 & 56.1 & 76.5 & 87.8 & 52.0 & 40.0 & 30.0 & 60.0 \\
 & Action counts & 10.4 & 11.4 & 11.1 & 4.9 & 10.9 & 5.5 & 5.3 & 3.8 \\
\midrule
\multirow{2}{*}{LLaMA-3.2-3B}
 & Final label   & 87.5 & 71.0 & 76.5 & 91.1 & 60.0 & 52.0 & 35.0 & 40.0 \\
 & Action counts & 10.6 & 9.2 & 11.2 & 4.2 & 10.0 & 5.1 & 5.2 & 4.0 \\
\midrule
\multirow{2}{*}{LLaMA-2-Chat-7B}
 & Final label   & 87.5 & 64.0 & 80.5 & 93.3 & 32.0 & 60.0 & 45.0 & 80.0 \\
 & Action counts & 10.9 & 10.0 & 9.7 & 3.9 & 11.4 & 5.2 & 5.3 & 4.0 \\
\midrule
\multirow{2}{*}{LLaMA-3.1-8B}
 & Final label   & 86.5 & 73.0 & 85.5 & 94.4 & 68.0 & 40.0 & 35.0 & 75.0 \\
 & Action counts & 11.0 & 8.9 & 12.5 & 3.8 & 9.7 & 5.4 & 5.1 & 4.0 \\
\midrule
\multirow{2}{*}{Qwen-3-0.6B}
 & Final label   & 88.5 & 49.0 & 75.5 & 91.1 & 60.0 & 12.0 & 25.0 & 55.0 \\
 & Action counts & 10.6 & 12.1 & 11.2 & 4.5 & 10.2 & 5.8 & 5.5 & 4.0 \\
\midrule
\multirow{2}{*}{Qwen-3-1.7B}
 & Final label   & 90.0 & 62.0 & 75.0 & 91.1 & 44.0 & 36.0 & 40.0 & 60.0 \\
 & Action counts & 10.5 & 10.3 & 11.2 & 4.5 & 11.2 & 5.3 & 5.6 & 4.2 \\
\midrule
\multirow{2}{*}{Qwen-3-4B}
 & Final label   & 84.5 & 68.0 & 75.5 & 91.1 & 80.0 & 52.0 & 40.0 & 75.0 \\
 & Action counts & 11.2 & 9.7 & 11.3 & 4.1 & 9.0 & 5.0 & 5.2 & 4.1 \\
\bottomrule
\end{tabular}
\end{table}

\subsection{Component Ablation Performance}
Table \ref{app::abla_compoent_tab} presents detailed results for the component ablation study across all environments. Consistent with Section \ref{sec:contextcomp}, removing action history, available actions, or observation history leads to environment-specific performance drops, highlighting the differential importance of each context component. For instance, masking action history or available actions significantly reduces performance in ALFWorld and Wordle, while omitting observation history most strongly affects TextCraft. The full belief model with all components retained consistently achieves the best balance of task completion and action efficiency, reinforcing the value of comprehensive context modeling in \gls{pabu}.

\begin{table}
\centering
\caption{Component × Environment Results}
\label{app::abla_compoent_tab}
\setlength{\tabcolsep}{5pt}
\begin{tabular}{l|l|rrrrrrrr}
\toprule
\textbf{Model} & \textbf{Metric} & \textbf{ALF} & \textbf{TC} & \textbf{Sci} & \textbf{Baby} & \textbf{MZ} & \textbf{WD} & \textbf{WT} & \textbf{MV} \\
\midrule
\multirow{2}{*}{No action history}
 & Final label   & 75.0 & 59.0 & 68.0 & 88.9 & 64.0 & 0.0 & 25.0 & 50.0  \\
 & Action counts & 13.1 & 10.6 & 12.8 & 4.8 & 9.4 & 6.0 & 5.2 & 4.1      \\
\midrule
\multirow{2}{*}{No available actions}
 & Final label   & 80.5 & 46.0 & 34.5 & 90.0 & 52.0 & 36.0 & 15.0 & 70.0 \\
 & Action counts & 13.3 & 12.9 & 21.3 & 4.3 & 11.5 & 5.4 & 5.2 & 4.1     \\
\midrule
\multirow{2}{*}{No observation history}
 & Final label   & 88.0 & 49.0 & 77.0 & 92.2 & 44.0 & 4.0 & 25.0 & 60.0 \\
 & Action counts & 10.4 & 12.3 & 10.7 & 4.0 & 11.8 & 5.9 & 5.2 & 3.9    \\
\midrule
\multirow{2}{*}{\gls{pabu} Full Belief}
 & Final label   & 89.5 & 60.0 & 76.5 & 87.8 & 52.0 & 40.0 & 35.0 & 65.0 \\
 & Action counts & 10.4 & 10.7 & 11.1 & 4.9 & 10.9 & 5.5 & 5.3 & 3.8     \\
\bottomrule
\end{tabular}
\end{table}

\section{More References on LLM Agent}
Agentic tasks represent one of the most demanding frontiers for LLM applications, as they require models to operate not just as passive responders but as autonomous decision-makers embedded within dynamic environments. Agentic applications demand continuous perception of the external state, integration of contextual signals, and formulation of multi-step action plans that unfold over time. This requires long-horizon reasoning, robust history management, and adaptability to feedback loops where each action reshapes the environment and constrains future options. 

\paragraph{Agentic training with optimized data pipelines.} 
In supervised fine-tuning (SFT), some works focus on instruction generation models. For example, \cite{sft_data_1} fine-tuned an instruction model using outputs from actor and reflection models powered by proprietary OpenAI APIs, while \cite{sft_data_7} trained an instruction sampling model from seed instructions and generated action sequences for downstream policy training. Other works \cite{sft_mcts_3, sft_data_3, sft_data_4, sft_data_0, sft_data_6, sft_data__1} developed alternative data collection pipelines for agentic tasks and applied SFT on small language models (SLMs) to behavior-clone full trajectories and acquire task-specific abilities.  

To reduce noise in otherwise successful trajectories, \cite{sft_data_5} used an LLM judge to filter actions, \cite{sft_data_8} generated and refined trajectories while excluding erroneous steps from the SFT loss, and \cite{sft_data_9} synthesized reflections based on action consistency to augment training data. Building on these approaches, \citep{sft_alg_1} proposed knowledge-based self-learning to evaluate and filter trajectories with a continuously updated memory, and \citep{sft_alg_2} introduced a self-iterative fine-tuning method on positive solutions, leveraging augmented correct steps to reflect on previous mistakes.

In reinforcement learning (RL), several works carefully design task-specific feedback for agentic training. For example, \citep{rl_data_2} curated preference pairs by synthesizing reasoning traces for ground-truth actions using proprietary models. Other approaches construct RL data through repeated sampling: ETO \citep{rl_data_0} proposed to use successful and failure trajectories for preference pairs in DPO, \citep{rl_data_3} applied MCTS with corrections at the first error step of a failed trajectory, and \citep{rl_data_4} dynamically adjusted the scope of action exploration and prioritized critical action steps for downstream DPO.

Some works adopt a sequential training pipeline combining SFT followed by RL. For example, \cite{both_seq_1} incorporated LLM API reflections into training, improving policy performance through both SFT and RL. \cite{both_seq_2} studied the optimal timing to start the RL phase, showing that early RL is beneficial but requires several initial epochs of SFT, with data generated from a 70B teacher model.  

Additionally, \cite{both_seq_3} introduced a top-layer planner that is first fine-tuned on seed plans and then optimized via DPO with interaction data. A third stage, Rejected Sampling Fine-Tuning (RFT), is applied by \cite{both_seq_4} after the SFT-then-RL pipeline to further refine the policy on high-reward actions.

For test-time scaling, \cite{sft_mcts_2} proposed mixing general and agentic data (inspired by \cite{sft_mcts_3}) and incorporating multi-path reasoning at inference time. Similarly, \cite{sft_mcts_1} fine-tuned the policy model on one quarter of the collected data, used the remaining three quarters to train the reward model, and found that explicit reward modeling with beam search yielded the best performance. 

\paragraph{Agentic training with novel algorithms.}
Several works have proposed novel algorithmic approaches for agentic tasks. \citep{rl_data_1} introduced a reward function for web shopping that evaluates both product and attribute selection. Building on reinforcement learning, \cite{rl_alg_1, rl_alg_6} proposed a hierarchical multi-turn RL framework that assigns rewards at the sentence or utterance level while updating policies at the token level, similar to DMPO \citep{rl_alg_3}, which introduced an RL-based step-wise alternative to DPO with discounting factors, and \cite{rl_alg_0} with a curriculum learning procedure with critic model scoring and modify the learning objective to relax the need of paired data. To better use multi-step reasoning, \cite{rl_alg_5} proposed OREO for maximum-entropy learning, eliminating the need for pairwise data and relying solely on a value function. Complementary approaches include \cite{rl_alg_4}, which learns stepwise rewards via inverse reinforcement learning (IRL) and tunes LLM agents with PPO, and \cite{rl_alg_2}, which calibrates rewards using L2 distances from demonstrations to mitigate reward hacking.

On the one hand, some reseraches jointly trained the policy model on a mixture of SFT and RL losses. \cite{both_sametime_1} combined SFT and RL loss with special masked loss with designed buffer structure, \cite{both_sametime_2} proposed a new objective weighted SFT by the learned reward, \cite{both_sametime_0} combined outcome-based and step-based DPO losses with SFT loss, and \cite{rl_alg_7} enhanced RL method with pseudocode planning, which incorporated SFT, DPO, and planning execution losses.

\section{Use of AI Assistants}
We utilized LLM-based tools solely for language refinement in manuscript preparation and for drafting icons in Fig. \ref{fig:motivation} (the final figure was manually produced in draw.io). All content generated by these models was thoroughly reviewed and verified by the authors, who take full responsibility for any errors or inaccuracies.

\end{document}